\documentclass{article}

\usepackage[final]{neurips_2023}

\usepackage[utf8]{inputenc} 
\usepackage[T1]{fontenc}    
\usepackage{hyperref}       
\usepackage{url}            
\usepackage{booktabs}       
\usepackage{amsfonts}       
\usepackage{nicefrac}       
\usepackage{microtype}      
\usepackage{xcolor}         

\title{FastCAD: Real-Time CAD Retrieval and Alignment from Scans and Videos}

\author{%
Florian Langer$^{1,2}$\thanks{This work was done as part of an internship at Qualcomm Technologies, Inc.} \quad Jihong Ju $^{1}$ \quad Georgi Dikov$^1$ \quad Gerhard Reitmayr$^1$ \quad Mohsen Ghafoorian$^1$ \\
$^1$XR Labs, Qualcomm Technologies, Inc. \quad $^2$Department of Engineering, University of Cambridge\\
\texttt{fml35@cam.ac.uk}\\
\texttt{\{jihoju,gdikov,gerhardr,mghafoor\}@qti.qualcomm.com}
}

\usepackage{graphicx}
\usepackage{booktabs}
\usepackage{bbm}
\usepackage{tikz}
\usepackage{comment}
\usepackage{amsmath,amssymb,amsbsy} 
\usepackage{color}
\usepackage{geometry}
\usepackage{multirow}
\usepackage{makecell}
\usepackage{colortbl}
\usepackage{appendix}
\usepackage{makecell}
\usepackage[accsupp]{axessibility}

\usepackage{hyperref}

\usepackage{orcidlink}

\begin{document}

\maketitle

\begin{abstract}


Digitising the 3D world into a clean, CAD model-based representation has important applications for augmented reality and robotics. Current state-of-the-art methods are computationally intensive as they individually encode each detected object and optimise CAD alignments in a second stage.
In this work, we propose FastCAD, a real-time method that simultaneously retrieves and aligns CAD models for all objects in a given scene.
In contrast to previous works, we directly predict alignment parameters and shape embeddings.
We achieve high-quality shape retrievals by learning CAD embeddings in a contrastive learning framework and distilling those into FastCAD.
Our single-stage method accelerates the inference time by a factor of 50 compared to other methods operating on RGB-D scans while outperforming them on the challenging Scan2CAD alignment benchmark.
Further, our approach collaborates seamlessly with online 3D reconstruction techniques. This enables the real-time generation of precise CAD model-based reconstructions from videos at 10 FPS. Doing so, we significantly improve the Scan2CAD alignment accuracy in the video setting from 43.0\% to 48.2\%
and the reconstruction accuracy from 22.9\% to 29.6\%.

\end{abstract}

\begin{figure*}[t]
    \centering
    \includegraphics[width=1.0\linewidth]{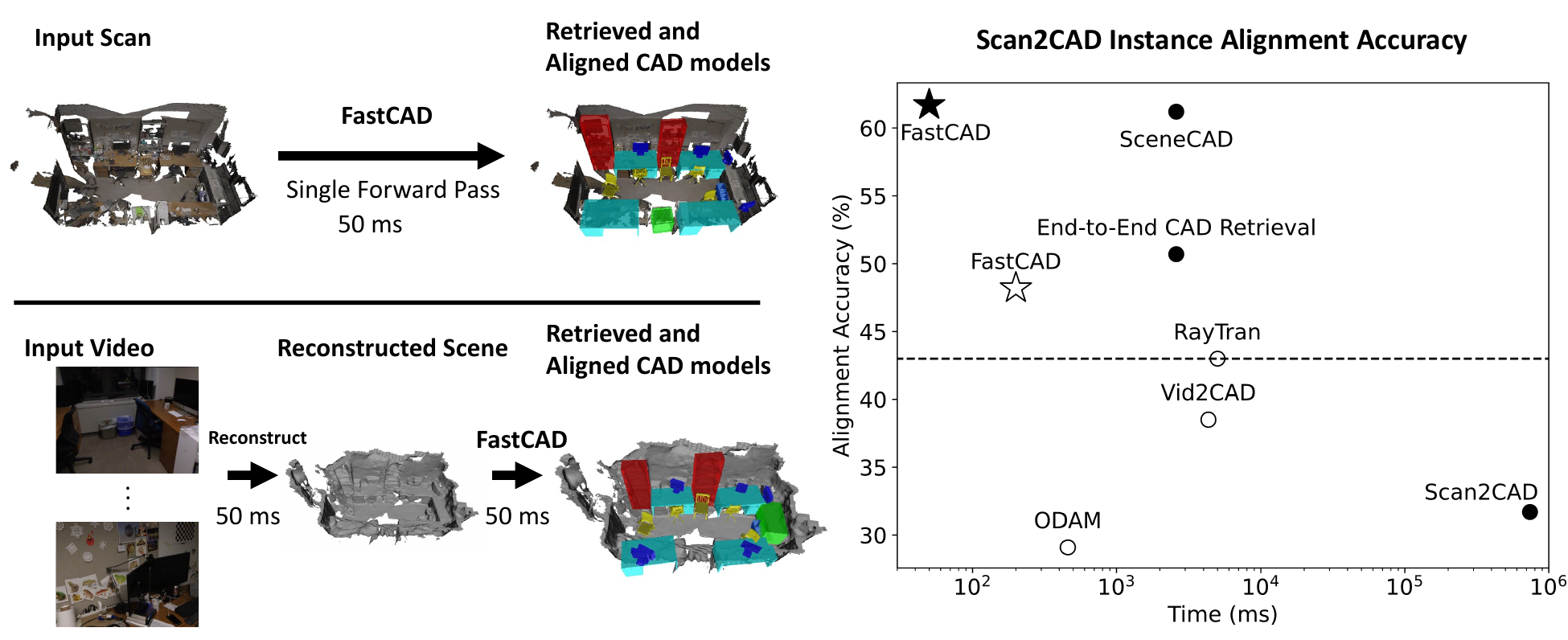}
    \caption{\textbf{Intro: FastCAD retrieves and aligns CAD models to point clouds in real-time.} FastCAD can either directly operate on an RGB-D scan (left top) or on the output of an off-the-shelf reconstruction method, which takes an RGB video as its input (left bottom). The graph on the right shows the Scan2CAD \cite{scan2cad} instance alignment accuracy as a function of inference time compared to competing methods. Note that the inference time is displayed on a log scale. Closed circles and stars (ours) denote methods operating on RGB-D scans, while open circles and stars represent methods using RGB videos as inputs. FastCAD outperforms previous methods in both settings while being significantly faster than the previously fastest methods. Note that RayTran \cite{raytran} did not disclose their run-times but is most likely much slower than FastCAD (see Supp. Mat.).}
    \label{fig_intro}
\end{figure*}

\section{Introduction}
Representing environments and rooms by aligned 3D CAD models is crucial for many downstream tasks in augmented reality or robotics. Compared to noisy 3D scene meshes or point clouds, a CAD-based representation has many advantages, such as the absence of holes in objects, clean surface geometry, object-level annotations, and potential part-level scene understanding. Additionally, the representation is more compact, with significantly fewer vertices and faces, which allows for faster rendering and collision simulations.\\
In this work, we introduce FastCAD, which is carefully designed to perform real-time CAD retrieval and alignment (see Fig. \ref{fig_intro}).
First, to achieve this goal, FastCAD
simultaneously solves object alignment and retrieval thanks to the proposed embedding distillation technique.
For this, we first learn an embedding space by training a separate encoder network in a contrastive learning setting. Noisy, partial scans and clean CAD models are embedded into a unified embedding space. By introducing two auxiliary tasks, performing foreground-background segmentation of the noisy object scan and predicting the similarity of the positive and negative CAD model used for the contrastive setup, we further improve the quality of the learned embeddings. Rather than using the encoder network to obtain embedding vectors at inference time, we distil the embeddings into FastCAD by supervising its shape embedding prediction per detection by the embedding of the ground-truth CAD model. Doing so greatly improves the speed as well as the quality of the retrieved shapes.

Second, FastCAD directly predicts alignment parameters.
We parameterise the alignments with oriented 3D bounding boxes where we additionally predict the front-facing side of the CAD model within the bounding box. This is significantly faster than analysis-by-synthesis-based methods \cite{scannotate,mcss} where CAD alignments are obtained iteratively by minimising rendering-based alignment objectives. It is also more efficient than correspondence-based methods \cite{scan2cad,end_to_end_cad_retrieval,scenecad} where the network outputs object-to-CAD correspondences and object poses are extracted with an additional alignment optimisation \cite{procrustes}.
At inference time, the shape embeddings predicted by FastCAD are used to retrieve the nearest neighbor CAD models from the embedding space. 
Those CAD models are aligned inside the predicted bounding boxes according to the predicted front-facing side to form the final output.
In this way, we achieve a very efficient method running in just 50 ms per RGB-D scan (compared to \cite{end_to_end_cad_retrieval,scenecad}, which takes 2.6 s) while achieving a similar accuracy on the Scan2CAD alignment benchmark compared to \cite{scenecad} (61.7\% vs 61.2\%, see Fig. \ref{fig_intro}).

Third, we can use FastCAD
in conjunction with reconstruction methods (e.g. \cite{simplerecon, DGrecon,transformerfusion}) to perform precise, real-time CAD alignments from videos. For this, we sample a point cloud from the output mesh generated with \cite{DGrecon} and use it as the input to FastCAD to predict CAD alignments.
Our results demonstrate that this way of first reconstructing an object-agnostic 3D scene representation and then performing object detection is more robust than frame-based methods \cite{odam,vid2cad}.
Further, choosing an explicit 3D point cloud as an intermediate representation means that 3D reconstruction methods can be used out-of-the-box and can be applied in an online setting, unlike \cite{raytran}.
Applying FastCAD on the output of \cite{DGrecon} our joint system can run online at 10 FPS (compared to less than 3 FPS \cite{odam}) while significantly improving the instance alignment accuracy on the Scan2CAD alignment benchmark from 43.0\% \cite{raytran} to 48.2\%.
Additionally, we introduce two metrics to assess the quality of retrieved shapes on the Scan2CAD \cite{scan2cad} benchmark and show that FastCAD improves the introduced reconstruction accuracy from 22.9\% \cite{vid2cad} to 29.6\%.
In summary, our key contributions include:
\begin{itemize}
    \item a novel and effective method for CAD model-based reconstruction where
    high-quality shape embeddings learned in a contrastive learning framework are distilled into an object detection network.
    \item an efficient system that predicts CAD retrievals and alignments for all objects in a scan in just 50 ms, allowing for online application to videos at 10 FPS.
    \item state-of-the-art alignment accuracy on the challenging and commonly used Scan2CAD benchmark for methods operating on scans (61.7\% vs 61.2\%) and on videos (48.2\% vs. 43.0\%).
    \item new evaluation metrics for the Scan2CAD benchmark assessing the quality of the retrieved shapes.
\end{itemize}

\section{Related Work}
Related work for this project comprises methods for \textit{CAD retrieval and alignment}, \textit{3D object detection} as well as general approaches for \textit{CAD retrieval from an embedding space}.

\subsection{CAD Retrieval and Alignment}
\textbf{Using RGB-D scans as inputs.}
Methods like \cite{end_to_end_cad_retrieval,scenecad} use predicted bounding boxes to crop parts of a feature volume, which are fed through a separate encoder to obtain shape embedding vectors. This is slower than our single-stage approach. To obtain CAD alignments \cite{scan2cad,end_to_end_cad_retrieval,scenecad} predict 3D correspondences for each object individually and then optimise for rotation and translation.
\cite{scenecad} additionally predicts scene-layout elements and refines the positions of the CAD models to obey support relations in their scene graph. Their run times range from ca. 20 minutes \cite{scan2cad} to 2.6 seconds \cite{end_to_end_cad_retrieval,scenecad}. Other methods \cite{mcss,scannotate} exhaustively render all CAD models in a database and optimise the pose of the best-fitting one by comparing rendered depth images to observed ones. However, with run times of more than 10 minutes per scene, these are not suited for real-time applications.\\
\textbf{Using RGB videos as inputs.}
\cite{odam,vid2cad,raytran} predict CAD alignments from posed RGB videos. \cite{odam,vid2cad} both individually detect objects in each frame, associate them across frames, and perform a multi-view optimisation to find the best pose for each object.
Approaches such as \cite{odam,vid2cad} are very engineered and, due to the heterogeneity of their different modules, can usually not be trained end-to-end. This fact, in combination with a brittle tracking-by-detection step, makes them error-prone and unreliable.
RayTran \cite{raytran} does not perform per-frame predictions
and instead relies on propagating the information into a 3D scene volume and performs predictions here.
While doing so, they avoid the issues mentioned above, their mechanism for creating a 3D feature volume is computationally expensive with undisclosed run times and, in its current form, can not be run in an online setting.

\subsection{3D Object Detection}
3D object detection methods can be grouped by their underlying mechanism of aggregating per-object information. \textit{Voting-based methods} \cite{votenet,brnet,h3dnet,rbgnet} are initialised with a set of candidate object centres and require points to vote on whether they belong to a given object. Features of all points that voted to be part of the same object are aggregated and decoded to obtain a bounding box prediction.
\textit{Attention-based methods} \cite{groupfree,3detr} impose fewer inductive biases than voting-based methods. They replace the voting-based method of determining which features to aggregate with an attention-based method, resulting in softer assignments and alleviating the need for some hyper-parameters.
\textit{Convolution-based methods} \cite{minkowski_2,fcaf3d,tr3d} convert point clouds into voxels and process them with 3D convolutions. Densely processing features in 3D is very memory and compute-intensive. GSDN \cite{minkowski_2} improves the efficiency for such 3D convolution-based methods by introducing a generative sparse tensor decoder using a series of transposed convolutions and pruning layers. FCAF3D \cite{fcaf3d} used those transposed convolutions but introduced an anchor-free method that can better model the diversity of 3D object orientations and sizes.
Simplifying the network architecture of \cite{fcaf3d} and introducing a multi-level object assigner \cite{tr3d} achieves a run-time of 21 FPS while further improving the performance. FastCAD uses the same backbone and neck as \cite{tr3d}.
\subsection{CAD Retrieval from an Embedding Space}
Various previous works \cite{shrec,joint_embedding} have investigated learning an embedding space from which CAD models can be retrieved to model real-world objects. The most relevant of such works is \cite{joint_embedding}, which learns a joint embedding space of noisy, incomplete scan objects and clean CAD models. They use 3D convolutions to learn feature embeddings for scans and CAD models. Their convolutional layers are trained by minimising a triplet loss \cite{triplet} where they sample CAD models of a different category as negatives. Other works such as \cite{image_purification,mask2cad,patch2cad,leveraging_geometry} learn CAD model embedding spaces by rendering CAD models and learning embeddings for the rendered and real images in a contrastive learning setting.
\section{Method}
\label{sec_method}

\begin{figure*}[t]
    \centering
    \includegraphics[width=1.0\linewidth]{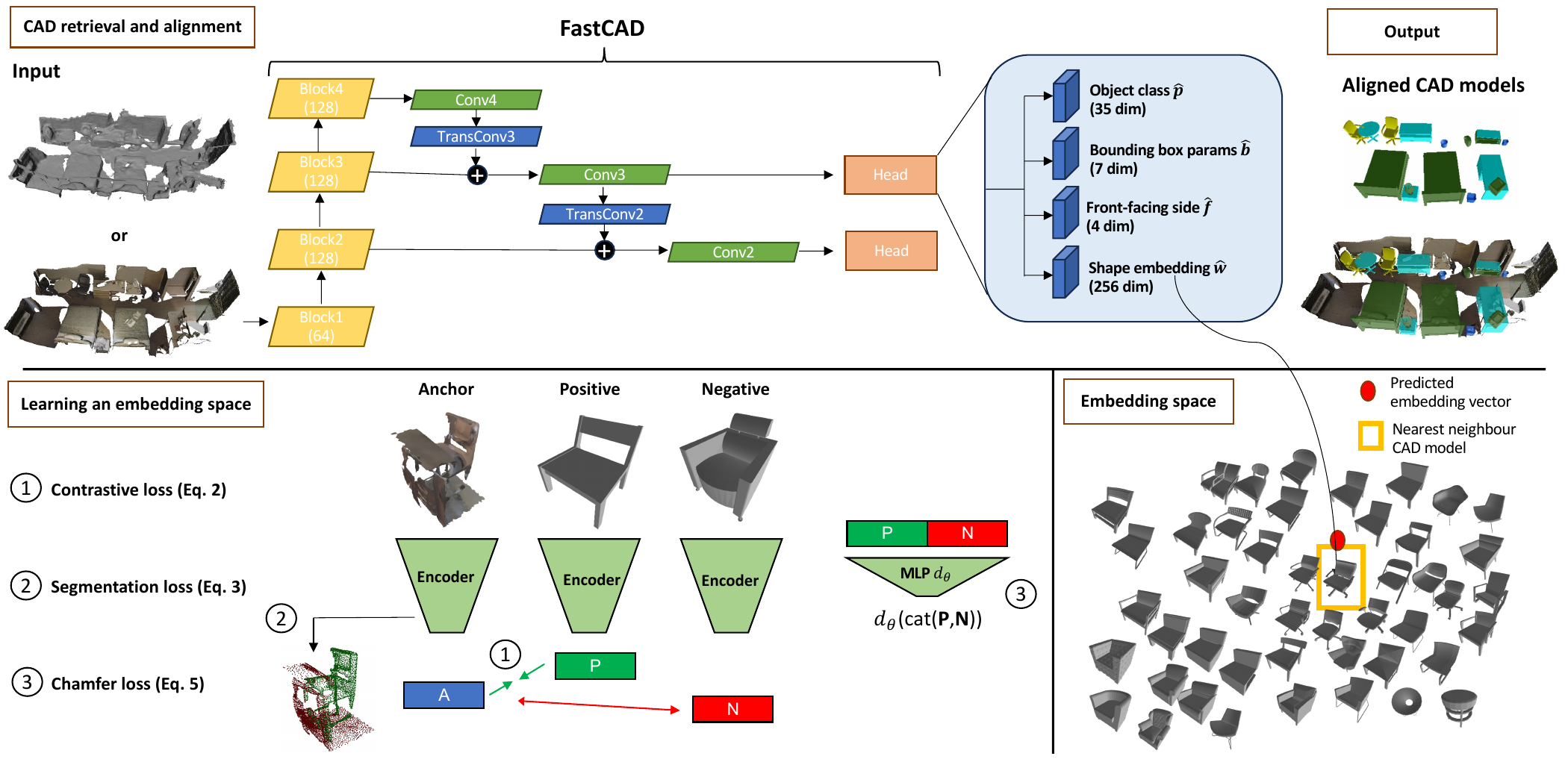}
    \caption{\textbf{Method.} FastCAD retrieves and aligns CAD models for all objects detected in an input point cloud. For all detected objects it predicts their category $\hat{\boldsymbol{p}}$, bounding box parameters $\hat{\boldsymbol{b}}$, front-facing side $\hat{\boldsymbol{f}}$ and shape embedding $\hat{\boldsymbol{w}}$. The predicted embedding vector $\hat{\boldsymbol{w}}$ is used to retrieve the nearest neighbour CAD model from an embedding space previously learned in a contrastive learning setting with auxiliary tasks.} 
    \label{fig_method}
\end{figure*}
FastCAD (Fig. \ref{fig_method}) simultaneously predicts CAD alignments and shape embeddings for objects detected in a point cloud (Sec. \ref{sec_cad_retrieval_and_align}). The predicted shape embeddings are used to retrieve the nearest CAD models from an embedding space. This embedding space is learned by encoding noisy, partial object scans and clean CAD models into a joint embedding space in a contrastive learning setting (Sec. \ref{sec_embedding_space}).


\subsection{CAD Retrieval and Alignment}
\label{sec_cad_retrieval_and_align}

The input to FastCAD is a point cloud, which may be derived from (i) an RGB-D scan or (ii) a noisy scene reconstruction obtained, for example, by applying \cite{simplerecon, DGrecon,transformerfusion} to a video.
This point cloud is encoded into a feature volume using a set of sparse 3D convolutions followed by generative transposed convolutions \cite{gsdn}. FastCAD's network architecture is inspired by \cite{tr3d}. 
For a range of sampled locations $(\hat{x},\hat{y},\hat{z})$ a shared detection head outputs classification probabilities $\hat{\boldsymbol{p}}$, oriented bounding box parameters $\hat{\boldsymbol{b}}$, front-facing side classification $\hat{\boldsymbol{f}}$ and shape embedding vector $\hat{\boldsymbol{w}}$. Depending on the average size of the predicted object class, the head output at feature level 2 or 3 is returned (level 2 for small objects, level 3 for large objects).
For each oriented bounding box prediction $\hat{\boldsymbol{b}}$ we classify which of the four faces is the front face of the object using $\hat{\boldsymbol{f}}$. This information is used to choose between the four possible orientations when aligning the CAD model within the oriented bounding box. Encoding this information separately from the orientation in $\hat{\boldsymbol{b}}$ allows us to more easily leverage 
the symmetry annotations from Scan2CAD \cite{scan2cad} which label each object to be non-symmetric or have 2-fold, 4-fold or complete rotational symmetry around the up-axis. For 2-fold, 4-fold and complete rotational symmetric objects, we modify the target front-facing side $\boldsymbol{f}$ from, e.g. $\left(1,0,0,0\right)$ to $\left(\frac{1}{2},0,\frac{1}{2},0\right)$, $\left(\frac{1}{4},\frac{1}{4},\frac{1}{4},\frac{1}{4}\right)$ and $\left(\frac{1}{4},\frac{1}{4},\frac{1}{4},\frac{1}{4}\right)$ respectively. This prevents the network from overfitting to arbitrary orientations for symmetric objects and allows it to generalise better (see Tab. \ref{table_scan2cad_alignment_results}). 
Through an assignment procedure, a detection $i$ at $(\hat{x},\hat{y},\hat{z})$ may be matched with the nearest ground-truth object. This location then has ground-truth labels associated with it, and one can formulate a loss function as:
\begin{equation}
\mathcal{L}_{\mathrm{tot}}=\frac{1}{N_{\mathrm{mat}}} \sum_{i=1}^{N_{\text{det}}}
\mathcal{L}_{\mathrm{cls}}({\hat{\boldsymbol{p}}}_i, \boldsymbol{p}_i)
+ 
\mathbbm{1}_{i}
\left(
\mathcal{L}_{\mathrm{bb}}({\hat{\boldsymbol{b}}}_i, \boldsymbol{b}_i)
+ \mathcal{L}_{\mathrm{ff}}({\hat{\boldsymbol{f}}}_i, \boldsymbol{f}_i)+ \mathcal{L}_{\mathrm{emb}}({\hat{\boldsymbol{w}}}_i, \boldsymbol{w}_i)
\right)
\end{equation}
For each loss, predicted values are denoted with a hat.
$\mathbbm{1}_{i} = 1$ if detection $i$ is matched to a ground-truth object and $\mathbbm{1}_{i} = 0$ if not.
$N_{\mathrm{mat}} = \sum_{i=1}^{N_{\text{det}}}
\mathbbm{1}_{i}$ is the total number of matches. If a detection is not matched to a ground-truth object $\boldsymbol{p}_i = \boldsymbol{0}$. 
Note that each detection can be matched to only one ground-truth object, which is only matched if it is among the $k=6$ closest detections to that ground-truth object.
The classification loss $\mathcal{L}_{\mathrm{cls}}$ is a focal loss, the bounding box loss $\mathcal{L}_{\mathrm{bb}}$ is a DIoU loss \cite{diou}, the front-facing side loss $\mathcal{L}_{\mathrm{ff}}$ is a cross-entropy loss and the shape loss $\mathcal{L}_{\mathrm{emb}}$ is a MSE loss.
To obtain ground-truth shape embedding vectors $\boldsymbol{w}_i$ we first learn a CAD model embedding space (see Sec. \ref{sec_embedding_space}).

\subsection{Learned Embedding Space}
\label{sec_embedding_space}
We learn a shape embedding space using a contrastive learning setup with two new auxiliary tasks.
For contrastive learning, we embed noisy object point clouds from scans and clean point clouds sampled from CAD models into a unified embedding space. For this purpose, we select all points within the Scan2CAD \cite{scan2cad} object bounding boxes as anchor objects and associate the point clouds of the annotated CAD model as the positive example. We randomly sample different CAD models of the same category as negative examples. These three point clouds are passed through an encoder network to produce embedding vectors $\boldsymbol{w}$. We employ a triplet loss \cite{triplet}:
\begin{equation}
\label{eq_triplet}
    \mathcal{L}_\mathrm{Contrastive} = \mathrm{max} (0,d^2(\textbf{A},\textbf{P}) + m - d^2(\textbf{A},\textbf{N})),
\end{equation}
where $\textbf{A}$, $\textbf{P}$ and $\textbf{N}$ are the embeddings of the anchor, positive and negative examples respectively. $d(\textbf{A},\textbf{B})$ denotes the L2 distance between vector $\textbf{A}$ and $\textbf{B}$.
This loss ensures that the distance between the anchor and the positive example is smaller by a margin $m$ than the distance between the anchor and the negative. 
In addition to the contrastive loss, we train the encoder to perform two auxiliary tasks. Doing so improves the quality of the retrieved shapes in FastCAD (see Tab. \ref{table_ablations}). The first task is to perform \textit{foreground/background segmentation} of the input point clouds of the real scan. This is supervised with a binary cross-entropy loss:
\begin{equation}
    \mathcal{L}_\mathrm{Segmentation}=-\frac{1}{N_{\mathrm{Seg}}} \sum_{i=1}^{N_{\text{Seg}}} \left( y_i \text{log} (x_i) + (1-y_i) \text{log} (1-x_i)\right)
\end{equation}
Here $x_i \in [0,1]$ are the predicted probabilities for each point, $y_i \in \{0,1\}$ are the foreground/background labels and $N_{\text{Seg}}$ is the number of points sampled. Note that we balance the ratio of foreground to background labels by only applying a loss to as many foreground points as there are background points. Otherwise, ca. 80\%-90\% of sampled points belong to the foreground class and we observe slightly smaller improvements to the quality of the embeddings.

For the second task, we train a shallow MLP, $d_\theta$, to \textit{regress the Chamfer distance} between the positive and the negative CAD model from their embeddings. The Chamfer distance $\mathrm{d}_{\text {Chamfer}}(X, Y)$ for point clouds $X$ and $Y$ is defined as
\begin{equation}
\label{eq_chamfer_distance}
\mathrm{d}_{\text {Chamfer}}(X, Y)=\frac{1}{2} \left (\frac{1}{|X|} \sum_{x \in X} \min _{y \in Y} d(x, y) + \frac{1}{|Y|} \sum_{y \in Y} \min _{x \in X} d(x, y)\right )
\end{equation}
The introduced loss is 
\begin{equation}
    \mathcal{L}_\mathrm{Chamfer}=\left | \left |
    \mathrm{d}_\theta(\text{cat}(\textbf{P}, \textbf{N})) - \mathrm{d}_{\text {Chamfer}}(X_{\text{pos}}, X_{\text{neg}})
    \right | \right |_1,
\end{equation}
where $\mathrm{d}_\theta(\text{cat}(\textbf{P}, \textbf{N}))$ is the Chamfer distance predicted from the concatenated embeddings $\textbf{P}$ and $\textbf{N}$ of the positive and negative CAD model. $\mathrm{d}_{\text{Chamfer}}(X_{\text{pos}}, X_{\text{neg}})$ is the ground-truth Chamfer distance computed using Eq. \ref{eq_chamfer_distance}.
The intuition behind introducing this loss is that sometimes the negative CAD model can be similar to the positive CAD model, while at other times, it may be very different. Forcing the encoder network to learn embeddings containing such information helps learn more useful embeddings.
After training the encoder network, we compute embeddings for all CAD models in our training data.
At inference time for a given object detection and associated embedding prediction $\hat{\boldsymbol{w}}$ we retrieve the nearest neighbour CAD model of the predicted category $\hat{\boldsymbol{p}}$ and align it using the predicted bounding box $\hat{\boldsymbol{b}}$ and front-facing classification $\hat{\boldsymbol{f}}$.
\section{Experimental Setup}
\label{sec_experimental_setup}

\begin{figure*}[t]
    \centering
    \includegraphics[width=1.0\linewidth]{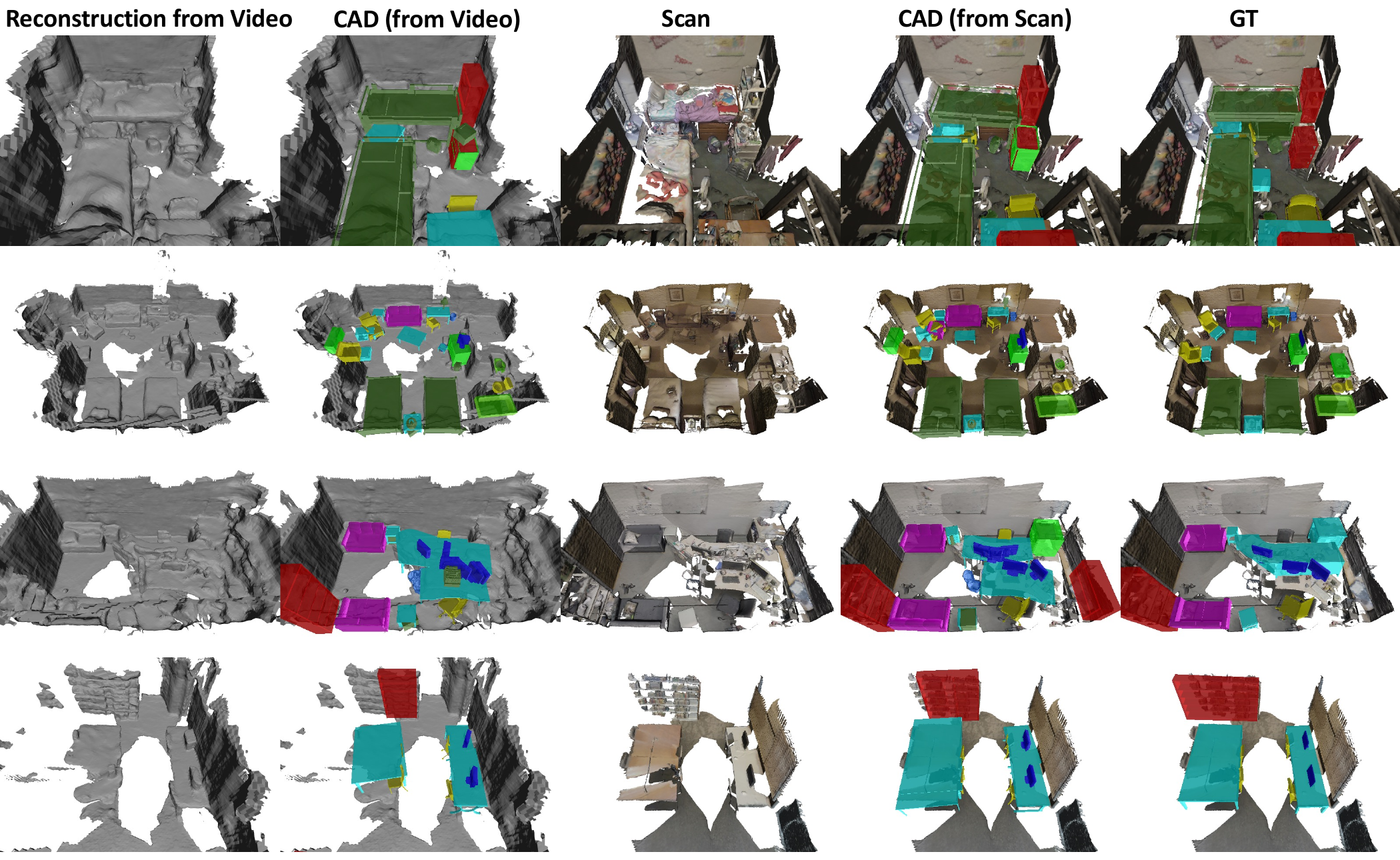}
    \caption{\textbf{Qualitative visualisation on ScanNet \cite{scannet,scan2cad}}. Column 1 shows the reconstruction generated by applying \cite{DGrecon} to the input video. Column 2 shows the CAD retrieval and alignments predicted by FastCAD when operating on the reconstruction in column 1. Columns 3 and 4 show the input scan from ScanNet \cite{scannet} and the CAD alignments FastCAD predicts for it. Column 5 shows the ground-truth CAD alignments from Scan2CAD \cite{scan2cad}.}
    \label{fig_qualitative}
\end{figure*}

\subsection{Dataset}
For training and testing our method, we use ScanNet \cite{scannet} with CAD model annotations provided by Scan2CAD \cite{scan2cad}. Those labels annotate the 1201 train scenes and 312 validation scenes from ScanNet \cite{scannet} with CAD models from ShapeNet \cite{shapenet}. There are over 14K objects annotated with over 3K unique CAD models, which come from 35 categories, with the most popular categories being chair, table and cabinet.
\subsection{Evaluation Metrics}
For evaluating the CAD alignments, we follow the original evaluation protocol introduced by Scan2CAD \cite{scan2cad}.
A CAD model prediction is considered correct
if the object class prediction is correct, the translation error is less than 20 cm, the rotation
error is less than 20°, and the scale error is below 20\%. 
For each scene and each category, as many predictions can be made as there are ground-truth CAD models. No duplicate predictions for the same ground-truth CAD model are considered.\\
\textbf{Introducing reconstruction and shape accuracy metrics.}
The metric above does not evaluate the quality of the aligned shapes. To do so, we introduce the
\textit{Scan2CAD reconstruction accuracy}. For this metric, the individual checks on rotation, translation and scale are replaced by checking if the F-score at $\tau$ between the aligned predicted and aligned target CAD model is larger than a threshold $\mu$ and consider those a correct prediction.
The F-score is defined as the harmonic mean of precision and recall, where precision is the fraction of points sampled on the predicted CAD model that lie within $\tau$ of points sampled on the ground-truth CAD model. Recall is the fraction of points on the ground-truth CAD within $\tau$ from a point on the predicted CAD model. Following \cite{meshrcnn}, before computing the F-scores, objects are rescaled such that the largest side of the ground-truth CAD model has a length of 10 so that small and large objects are compared fairly. We set $\tau = 0.5$ and $\mu=0.7$ (see the Supp. Mat. for results for different thresholds of $\mu$). We also introduce the \textit{Scan2CAD shape accuracy}, which follows the same protocol as the Scan2CAD reconstruction accuracy but computes the F-score when both the ground-truth and predicted CAD model are perfectly aligned, such as only focusing on the quality of the retrieved shape.

\subsection{Hyperparameters}
For training the encoder network, we process all CAD models by normalising them to a unit cube and randomly sampling 1024 points from their surface. Similarly, cropped object scans are normalised and 1024 points are randomly sampled. Point clouds from cropped object scans with less than 1024 points are padded with 0s. We apply random scaling between 0.9 and 1.1, random translation between -0.1 and 0.1 and random rotation between $-10^{\circ}$ and $10^{\circ}$ on all point clouds.
We use a Perceiver \cite{perceiver} as the encoder for the main experiments. It consists of three layers of cross-attention, each followed by two layers of self-attention, which share weights. The number of latent variables in the Perceiver \cite{perceiver} and their dimension is set to 256. The encoder network is trained for 750 epochs using a Lamb Optimiser \cite{lamb} with a learning rate of 1e-3 and batch size of 25. The learned shape embeddings $\boldsymbol{w}$ also have dimension 256. The margin $m$ in the triplet loss in Eq. \ref{eq_triplet} is set to 0.1. Foreground/background segmentation labels are predicted by cross-attending the final latent variables with the input point cloud. The MLP $d_\theta$ for predicting the Chamfer distance has a single hidden layer of size 256 and uses ReLU activation functions.\\
FastCAD is trained for 225 epochs using an AdamW optimiser \cite{adamw} with a learning rate set to 1e-3 and weight decay by a factor of 10 after 120 and 165 epochs. Before processing each scene, the corresponding point cloud is down-sampled to a maximum of 50,000 points.
During training, we perform a random sampling of input points (between 33\% and 100\%), random flipping along the x and y-axis with probability 50\% as well as random rotation around the z-axis (from $-\pi$ to $\pi$), random scaling (between 0.9 and 1.1) and random translation (between -0.5 m and 0.5 m).
Note that for predicting CAD alignments from videos, we train a separate version of FastCAD on the outputs of \cite{DGrecon} for the training scenes in ScanNet \cite{scannet}. This is because these more closely match the inputs that FastCAD receives at inference time for this setting.
\subsection{Implementation Details}
All code is implemented in PyTorch. 
FastCAD is integrated within the open-source object detection toolbox MMDetection3D \cite{mmdet3d2020}. It uses sparse convolutions from the Minkowski Engine \cite{minkowski_1,minkowski_2}. Training on a single RTX 2080 takes $\sim$7 hours. Training the encoder network on the same GPU takes $\sim$24 hours.

\section{Results}
In Sec. \ref{sec_results_alignments} we evaluate our CAD alignments on the Scan2CAD alignment benchmark. In Sec. \ref{sec_results_reconstruction_and_shape} we analyse the quality of the achieved reconstructions and the shape retrievals. Sec. \ref{sec_incremental_evaluation} presents results when evaluating FastCAD's predictions continuously while reconstructing a scene. Finally, Sec. \ref{sec_ablations} ablates various components of the proposed system.

\subsection{CAD Model Alignments}
\label{sec_results_alignments}

\begin{table}[t]
    \centering
    \resizebox{\textwidth}{!}{
    \begin{tabular}{l|ccccccccc|cc|c}
     Method & bathtub & bkshlf & cabinet & chair & display & sofa & other & table & trash bin & \textbf{class} & \textbf{instance} & time [ms]\\
     \hline
     Number of test instances \# & 120 & 212 & 260 & 1093 & 191 & 113 & 410 & 553 & 232 & 35 & 3184 & - \\
    \hline
    \multicolumn{13}{c}{\textbf{Competing Methods - Input RGB-D Scan}}\\
    \hline
    Scan2CAD \cite{scan2cad} & 36.2 & 36.4 & 34.0 & 44.3 & 17.9 & 30.7 & \textbf{70.6} & 30.1 & 20.6 & 35.6 & 31.7 & 740000 \\
    End-to-End CAD Retrieval \cite{end_to_end_cad_retrieval} & 38.9 & 41.5 & 51.5 & 73.0 & 26.5 & 76.9 & 26.8 & 48.2 & 18.2 & 44.6 & 50.7 & 2600 \\
    SceneCAD \cite{scenecad} & 42.4 & 36.8 & \textbf{58.3} & 81.2 & \textbf{50.7} & \textbf{82.9} & 40.2 & 45.6 & 32.3 & 52.3 & 61.2 & 2600 \\
    Ours (Scan) & \textbf{43.3} & \textbf{47.2} & 46.5 & \textbf{85.7} & 24.1 & 61.9 & 40.5 & \textbf{56.1} & \textbf{69.8} & \textbf{52.8} & \textbf{61.7} & \textbf{50} \\
    \hline
    \multicolumn{13}{c}{\textbf{Competing Methods - Input RGB Video}}\\
    \hline
    ODAM \cite{odam} & 24.2 & 12.3 & 13.1 & 42.8 & 36.6 & 28.3 & 0.0 & 31.1 & 42.2 & 25.6 & 29.2 & 366\\
    Vid2CAD \cite{vid2cad} & 28.3 & 12.3 & 23.8 & 64.6 & \textbf{37.7} & 26.5 & 6.6 & 28.9 & 47.8 & 30.7 & 38.6 & 3200\\
    RayTran \cite{raytran} & 19.2 & \textbf{34.4} & \textbf{36.2} & 59.3 & 30.4 & 44.2 & \textbf{27.8} & 42.5 & 31.5 & 36.2 & 43.0 & -\\
    Ours (Video) & \textbf{35.0} & 31.1 & 35.0 & \textbf{71.5} & 4.2 & \textbf{54.0} & 25.1 & \textbf{48.8} & \textbf{48.7} & \textbf{39.3} & \textbf{48.2} & \textbf{100}\\
    \hline
    \multicolumn{13}{c}{\textbf{Ablations - Front-Facing Side Prediction}}\\
    \hline
     Ours – discrete CAD orientation in embedding & 27.5 & 36.3 & 42.7 & 85.5 & \textbf{24.6} & 61.1 & 33.4 & 47.7 & 50.0 & 45.4 & 56.2 & 50 \\
    Ours – front-facing side prediction & 41.7 & 45.3 & 46.2 & 84.6 & 17.8 & 58.4 & 38.0 & \textbf{56.8} & 65.1 & 50.4 & 60.1 & 50 \\
    Ours – front-facing side prediction + symmetry & \textbf{43.3} & \textbf{47.2} & \textbf{46.5} & \textbf{85.7} & 24.1 & \textbf{61.9} & \textbf{40.5} & 56.1 & \textbf{69.8} & \textbf{52.8} & \textbf{61.7} & 50 \\
    \hline
    \end{tabular}}
    \caption{\textbf{Alignment accuracy on Scan2CAD}     
    \cite{scan2cad} in comparison to the state-of-the-art. All numbers (except time and the first row) are percentages, and higher is better. FastCAD outperforms competing methods on scans and videos while dramatically reducing the inference time in both cases.} 
    \label{table_scan2cad_alignment_results}
\end{table}
\begin{figure*}[t]
    \centering
    \includegraphics[width=1.0\linewidth]{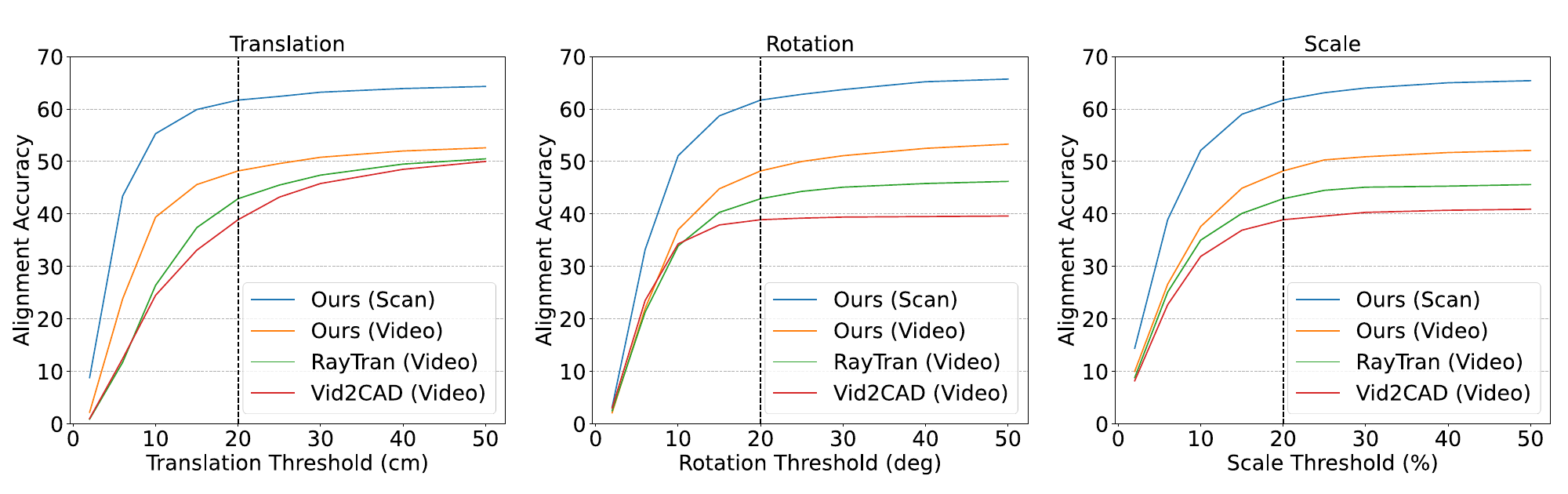}
    \caption{\textbf{Investigating different thresholds for the instance alignment accuracy on Scan2CAD \cite{scan2cad}}. The translation, rotation and scale thresholds, used to determine whether an alignment is correct, are varied from their default values at 20 cm, 20° and 20\%. Note that in each plot, the thresholds that are not investigated remain at their default value. FastCAD outperforms competing methods across all thresholds.}
    \label{fig_thresholds}
\end{figure*}
Comparing FastCAD to existing methods operating on RGB-D scans, we find that FastCAD performs similarly to SceneCAD \cite{scenecad}, the previously most accurate method (61.7\% vs. 61.2\% instance alignment accuracy) while being more than 50 times faster (see Tab. \ref{table_scan2cad_alignment_results}). This massive speedup is mainly due to direct prediction of CAD alignments and shape embeddings in a single step.
Compared to other methods when the input is an RGB video, FastCAD is not just significantly faster but also considerably more accurate, outperforming the following best-competing method RayTran \cite{raytran} by a large margin ($48.2\%$ vs. $43.0\%$ alignment accuracy). We also compare our alignments to previous works at different thresholds for computing the alignment accuracy (see Fig. \ref{fig_thresholds}). Here, we find that we outperform them across all settings.
Regarding run times, our total run-time to integrate new information from a new frame is just 100 ms (50 ms to run \cite{DGrecon} plus 50 ms to run FastCAD on the reconstructed scene). This is significantly faster compared to ODAM \cite{odam} (366 ms), Vid2CAD \cite{vid2cad} (3200 ms) and most likely also RayTran\footnote{See the Supp. Mat. for a discussion of this.} \cite{raytran} (see Fig \ref{fig_intro}).

We ablate our design decision for predicting the front-facing side $\hat{\boldsymbol{f}}$. In the first row of the last section in Tab. \ref{table_scan2cad_alignment_results} we present the accuracies when encoding the information about the front-facing side of a CAD model in the shape embedding $\boldsymbol{w}$. In this case, each CAD model has four embedding vectors for each of the four discrete 90-degree orientations associated with it. At inference time the CAD model is aligned inside the predicted bounding box according to the discrete orientation of its nearest-neighbour embedding $\boldsymbol{w}$. 
The second row shows the accuracies when predicting the object front-facing side with an extra classification head (as explained in Sec. \ref{sec_method}). This significantly improves the alignment accuracy (60.1\% vs. 56.2\%) while reducing the number of CAD embeddings that need to be stored and searched by a factor of four compared to the previous row. Finally, the last row shows that the alignment accuracy is further improved if the symmetry of the CAD model is taken into account when learning to predict the front-facing side (61.7\% vs. 60.1\%).

\subsection{Reconstruction and Shape Quality}
\label{sec_results_reconstruction_and_shape}
\begin{figure*}[t]
    \centering
    \includegraphics[width=1.0\linewidth]{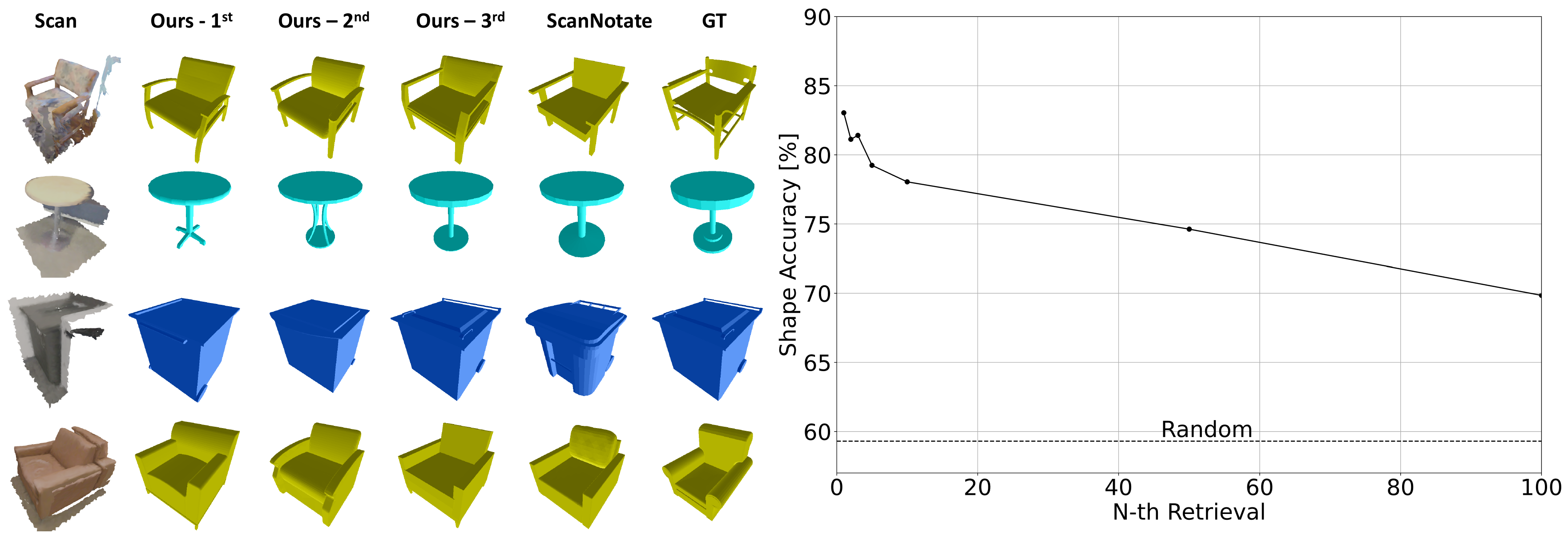}
    \caption{\textbf{CAD retrieval from the learned embedding space.} Left: Qualitative visualisation of the retrieved CAD model for a given object in a scene. Note that the input to FastCAD from which a shape embedding $\hat{\boldsymbol{w}}$ is predicted is the scan of the entire scene. However, for clearer visualisation, we only show the cropped part of the scan for which a CAD model is retrieved. Across different object categories, our CAD retrievals are of similar high quality as the ones from the pseudo-labelling method ScanNotate \cite{scannotate} and the ground-truth CAD models from Scan2CAD \cite{scan2cad}. Right: Our shape accuracy as a function of the N-th nearest CAD model retrieved from the embedding space. The shape accuracy remains high even as CAD models of increasingly worse rank are retrieved, which is a characteristic of a well-structured embedding space.}
    \label{fig_retrieval}
\end{figure*}

\begin{table}[t]
    \centering
    \resizebox{\textwidth}{!}{
    \begin{tabular}{l|l|c|c|c|c}
     &Method & Alignment Acc.  & Recon. Acc. & Shape Acc. & time [ms] \\
     \hline
     \multicolumn{6}{c}{\textbf{Competing Methods - Input RGB-D Scan}}\\
    \hline
     \multirow{2}{*}{Input RGB-D Scan} 
     & ScanNotate*  \cite{scannotate} & 78.2 & 60.1 & 83.5 & 660000 \\
     & Ours (Scan) & 61.7 & 41.7 & 83.1 & 50 \\
     \hline
     \multicolumn{6}{c}{\textbf{Competing Methods - Input RGB Video}}\\
    \hline
     \multirow{2}{*}{Input RGB Video}
     & Vid2CAD* \cite{vid2cad} & 38.6  & 22.9 & 76.6 & 3200\\
     & Ours (Video) & 48.2 & 24.7 & 79.8  & 100\\
     & Ours (Video, same retrieval Vid2CAD) &  48.2 & 29.6 & 87.7  & 100\\
    \hline
      \multicolumn{6}{c}{\textbf{Ablation Experiments -Input RGB-D Scan}}\\
    \hline
    \multirow{3}{*}{Embedding Distillation} & 2-step retrieval: pred bbox & 61.7 & 15.6 & 51.0 & 104 \\
    & 2-step retrieval: nearest GT bbox & 61.7 & 30.6 & 78.1 & 104 \\
    & Embedding distillation & 61.7 & 41.7 & 83.1 & 50 \\
    \hline
     \multirow{4}{*}{\makecell{Auxiliary Tasks for Training Encoder}}
    & Contrastive & 62.3 & 38.3 & 81.1 & 50 \\
    & Contrastive + Chamfer & 61.0 & 38.7 & 82.0 & 50\\
    & Contrastive + Segmentation & 61.3 & 41.5 & 84.3 & 50\\
    & Contrastive + Chamfer + Segmentation & 61.7 & 41.7 & 83.1 & 50 \\
    \hline
     \multirow{2}{*}{Encoder Architecture}
     & PointNet++ \cite{pointnet++} & 61.5 & 29.6 & 74.0 & 50\\
     & Perceiver \cite{perceiver} & 62.3 & 38.3 & 81.1 & 50 \\
    \hline
     \multirow{2}{*}{Different Input Sources}
     & ScanNet (Gray) & 60.4 & 40.1 & 82.9  & 50\\
     & DG Recon \cite{DGrecon} (Gray) & 48.2 & 24.7 & 79.8 & 100\\
    \hline
    \end{tabular}}
    \caption{\textbf{Alignment, reconstruction and shape accuracy on Scan2CAD \cite{scan2cad}} in comparison to competing methods and for various ablations. All accuracies are percentages and higher is better. Note that ScanNotate \cite{scannotate} initialises its CAD alignments from their ground-truth poses and Vid2CAD \cite{vid2cad} constraints its CAD retrieval to the very small ground-truth scene pool, making some of their results not exactly comparable to ours.}
    \label{table_ablations}
\end{table}
The CAD alignment accuracy used by \cite{scan2cad,end_to_end_cad_retrieval,scenecad,vid2cad,odam,raytran} does not evaluate the quality of the retrieved CAD models. We therefore introduce two metrics, the \textit{Scan2CAD reconstruction accuracy} and \textit{Scan2CAD shape accuracy} as explained in Sec. \ref{sec_experimental_setup}. While the \textit{Scan2CAD reconstruction accuracy} evaluates both the retrieved shapes and their alignments, the \textit{Scan2CAD shape accuracy} only evaluates the quality of the retrieved shapes.\\
\cite{scan2cad,end_to_end_cad_retrieval,scenecad,raytran} do not have publicly available code and were not able to share their shape retrievals with us.
We therefore compare our CAD retrievals to those from ScanNotate \cite{scannotate}. Note that ScanNotate \cite{scannotate} is used as an offline annotation method and optimises CAD retrievals and CAD poses, which are initialised from their ground-truth alignments\footnote{We exclude ScanNotate predictions for those objects that FastCAD did not detect to partially mitigate the effect of ScanNotate having access to perfect object detections.}. While the alignment and reconstruction accuracy of ScanNotate \cite{scannotate} is better than ours (because the objects are initialised from ground-truth poses), we find that the shape accuracy, focusing only on the quality of the retrieved CAD model but not their alignment, is similar to ours. This is a significant achievement given that \cite{scannotate} exhaustively renders all CAD models in the database, leading to run-times that are more than four orders of magnitude larger than ours.
In the video setting, FastCAD achieves better shape accuracies than Vid2CAD \cite{vid2cad} even when Vid2CAD \cite{vid2cad} limits its CAD retrievals to the ground-truth scene pool. When using the same retrieval setup as Vid2CAD \cite{vid2cad}, FastCAD achieves significantly better reconstruction accuracy ($29.6\%$ vs. $22.9\%$) and shape accuracy ($87.7\%$ vs. $76.6\%$).\\
To evaluate the quality of the embedding space, we compute the shape accuracy not just for the nearest neighbour retrieval, but also when retrieving instead the second, third or N-th nearest neighbour (see Fig \ref{fig_retrieval}). Here, we find that the shape accuracy remains high even when retrieving just the 10th closest CAD model. This demonstrates that geometrically similar CAD models are close to each other in the learned embedding space. This is desirable as it makes our CAD retrieval robust; even if the retrieved CAD model is not optimal, it will still closely match the observed object. Even retrieving the 100th closest CAD model from the learned embedding space is substantially more accurate than retrieving a random CAD model of the predicted category.
\subsection{Incremental Evaluation for Online Setting}
\label{sec_incremental_evaluation}
\begin{figure*}[t]
    \centering
    \includegraphics[width=1.0\linewidth]{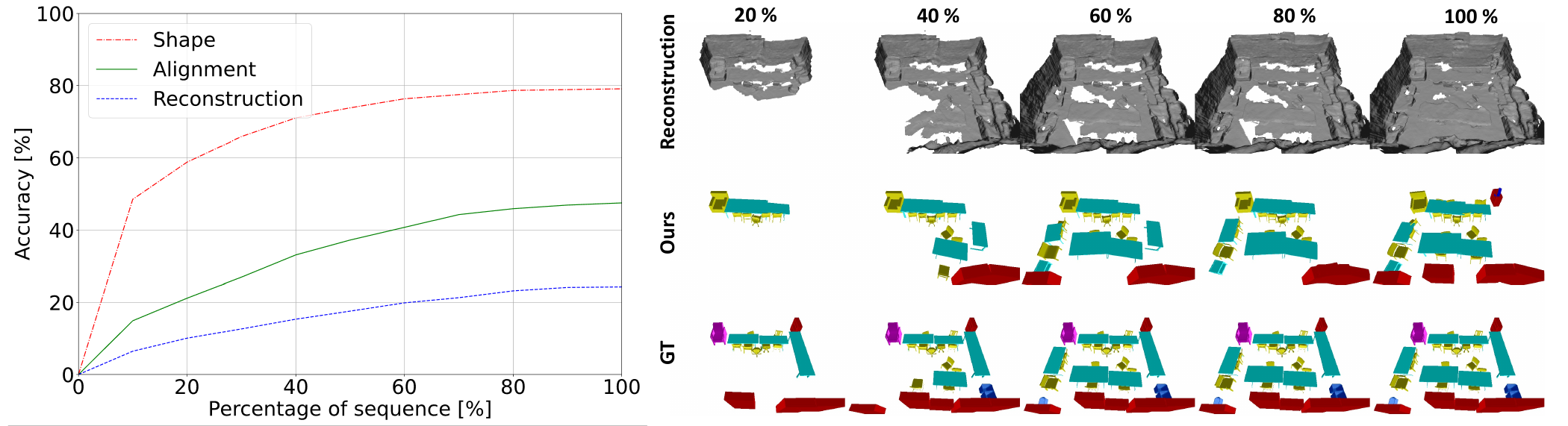}
    \caption{\textbf{Incremental evaluation of CAD predictions in an online setting.} Left: Various metrics are investigated when only parts of the RGB video sequence have been seen. Right: Visualisation of the incremental evaluation for one scene. The reconstructed scene mesh, our CAD alignments and the considered ground-truth CAD alignments are visualised at various stages of the RGB video sequence.}
    \label{fig_incremental}
\end{figure*}
When predicting CAD alignment from an RGB video, we can evaluate the different metrics at various stages of the video sequence (see Fig \ref{fig_incremental}). This is important for assessing the performance of our method for realistic applications in online settings, such as AR or robotics, where one requires not just an accurate final output but good performance throughout the sequence.
Note that for computing the metrics, only those ground-truth CAD models whose centre has already appeared in the field of view of at least one seen frame are considered. Here, we find that the investigated metrics show good performance even early on (ca. 30\% of the sequence). Nevertheless, we see a continuous improvement in all metrics as more parts of the video sequence are seen. We find that this is mainly because the output of the reconstruction method \cite{DGrecon} improves continuously as parts of the scene that previously appeared far away are observed from up close. This improved quality of the scene mesh leads to more accurate CAD predictions by FastCAD. Another reason for the improvements over time is simply occlusion. Some centres of ground-truth objects may have appeared in the field of view but have not been observed as they were hidden behind other objects. This means that \cite{DGrecon} can not reconstruct them, and consequently, FastCAD can not make predictions for those, reducing the accuracy.
\subsection{Ablations}
\label{sec_ablations}
We ablate various design choices of FastCAD in the lower part of Tab. \ref{table_ablations}. \\
\textbf{Embedding distillation.}
We investigate splitting bounding box detection and CAD retrieval into two successive steps. For this ablation, a detected bounding box is used to crop part of the input point cloud, which then serves as the input to the encoder to produce the shape embedding. This is the only experiment where the encoder is used at test time. We observe poor reconstruction and shape accuracy ($15.6\%$ and $51.0\%$). This is due to a distribution shift in the input to the encoder, which was trained on object point clouds cropped with ground-truth bounding boxes but now receives point clouds cropped with predicted bounding boxes. However, even when using the nearest ground-truth bounding box to crop the input for the encoder, the final reconstruction accuracy and shape accuracy are significantly worse compared to using FastCAD to directly predict shape embeddings ($41.7\%$ vs $30.6\%$ and $83.1\%$ vs. $78.1\%$). This demonstrates that FastCAD benefits from directly predicting shape embeddings as it can better integrate information from the surroundings and nearby objects.\\
\textbf{Auxiliary tasks for training encoder.}
We analyse the effect of training our encoder network with the two proposed auxiliary tasks. Here, we observe improvements in the reconstruction accuracy and shape accuracy for training by predicting the Chamfer distance between the positive and the negative CAD model as well as performing foreground/background classification of the input point cloud ($41.7\%$ vs $38.3\%$ and $83.1\%$ vs. $81.1\%$). These metrics are computed from FastCAD, which was trained to regress the shape embeddings but not directly trained with the additional losses. Better training of the encoder leads to improved embeddings, which, even after distilling those into FastCAD, leads to notably better reconstruction and shape accuracies.\\
\textbf{Encoder architecture.}
Testing different encoder architectures, we find that 
using a powerful encoder is crucial for obtaining high-quality shape embeddings. Compared to a standard PointNet++ \cite{pointnet++} network, using a Perceiver \cite{perceiver} increases the reconstruction accuracy from $29.6\%$ to $38.3\%$ and the shape accuracy from $74.0\%$ to $81.1\%$.\\
\textbf{Different input sources.}
The output of \cite{DGrecon} does not contain colour. To disentangle the effects of geometry and colour we input the point cloud from the RGB-D scan from ScanNet \cite{scannet} without any colour information. 
Comparing the alignment, reconstruction and shape accuracy, we observe that while the significantly noisier inputs affect the performance, the achieved outputs are still of high quality (see also Fig. \ref{fig_qualitative}). Comparing the experiments for the RGB-D scans without colour information to the main experiment, we also find that colour adds only very little information, and almost all information is contained in the geometry.\\  
\section{Conclusion}
We propose FastCAD, which can retrieve and align CAD models to an input scene scan in just 50 ms due to its efficient design. By applying FastCAD to the output of online 3D reconstruction techniques, we can obtain precise CAD-model-based reconstruction from videos running in real-time at 10 FPS. We train and validate our system on Scan2CAD \cite{scan2cad} which provides CAD model annotations for ScanNet \cite{scannet}.
Compared to competing works operating on scans, we reduce the run-time by a factor of 50 while slightly outperforming them regarding alignment accuracy. Compared to methods using videos as input, we improve the alignment accuracy from $43.0\%$ to $48.2\%$ while at least three times faster, thereby enabling real-time CAD-based reconstruction from videos. Future work could entail developing a mechanism to better ensure temporal consistency for FastCAD's retrievals and alignments in the online video setting.


\clearpage

\appendix
\appendixpage

\section{Run-Time Analysis of Competing Methods for RGB Videos}
Due to its efficient design, FastCAD can integrate information from a new frame into its CAD-based reconstruction in just 100 ms (50 ms for running \cite{DGrecon} and 50 ms for running FastCAD itself). 
ODAM \cite{odam} requires 166 ms for its object detection and object association and a further 200 ms to optimise each object pose. As different object poses can be optimised in parallel, their total time for integrating information from a new frame is 366 ms. Vid2CADs \cite{vid2cad} detection method takes 200 ms per frame, the tracking takes an additional 500 ms per frame, and the optimisation over poses takes 2.5 seconds, which gives it a run-time of 3200 ms.
RayTran \cite{raytran} does not provide any information in terms of run-time and the authors were not able to share any information regarding this with us. However, from their method design and their training requirements (8 x 16 GB GPU using 16-bit float arithmetic), one can infer that their method is extremely compute-intensive and can not integrate new information, except by running it again on all frames.

\section{Additional Results for the Reconstruction and Shape Accuracy}
Tab. \ref{table_supp_mat} shows the alignment, reconstruction and shape accuracy on Scan2CAD for different settings of $\mu$ (see Sec. 4.2 of the main paper for the definition of $\mu$). We find that, in general, the trends observed at $\mu=0.7$ can also be found at $\mu=0.5$ and $\mu=0.9$. In addition to providing results at extra settings for $\mu$, we include ablations for the number of input points in the last section of Tab. \ref{table_supp_mat}.
When reducing the number of input points from its default value of 50 K, we observe a graceful decline in the reconstruction accuracy and alignment accuracy. Note that even using 5 K points for the entire scene still yields good reconstruction and alignment accuracy.
We also find that the shape accuracy remains particularly high even for very low numbers of points. This means that FastCAD can still predict shape embeddings well in this data regime and the challenge lies more in accurate CAD alignment predictions.

\section{Poor Performance on Display Class}
When computing the CAD alignment accuracy per class in Tab. 1 in the main paper, we find that FastCAD performs significantly worse on the "display" class compared to other classes. Using scans as input, the alignment accuracy for displays is 24.1\% compared to the mean class accuracy of 52.8\%. Using videos as inputs, the "display" accuracy is just 4.2\% compared to the mean of 39.3\%. Visually inspecting the predictions we find that the issue in the majority of cases is a wrong prediction for the object front-facing side $\hat{\boldsymbol{f}}$. Investigating this phenomenon further we found that ShapeNet \cite{shapenet} CAD models of the "display" class are not oriented consistently. Out of 149 different "display" CAD models in the training set, 28 are facing the opposite direction. This results in a confusing training signal and explains the poor performance that is observed for this class. Ignoring this class would make our relative performance compared to competing methods even better.

\section{Further Visualisations}
\begin{table}[t]
    \centering
    \resizebox{\textwidth}{!}{
    \begin{tabular}{l|l|c|ccc|ccc|c}
     &Method & Alignment Acc. & \makecell{Recon. Acc.\\ $\mu=0.5$} & \makecell{Recon. Acc.\\ $\mu=0.7$} & \makecell{Recon. Acc.\\ $\mu=0.9$} & \makecell{Shape Acc.\\ $\mu=0.5$} & \makecell{Shape Acc.\\ $\mu=0.7$} & \makecell{Shape Acc.\\ $\mu=0.9$}  & time [ms] \\
     \hline
    \multicolumn{10}{c}{\textbf{Competing Methods - Input RGB-D Scan}}\\
    \hline
     \multirow{2}{*}{Input RGB-D Scan} 
     & ScanNotate*  \cite{scannotate} & 78.2 & 78.6 & 60.1 & 22.2 & 95.1 & 83.5 & 45.7 & 660000 \\
     & Ours (Scan) & 61.7 & 68.9 & 41.7 & 7.0 & 95.3 & 83.1 & 44.1 & 50 \\
     \hline
    \multicolumn{10}{c}{\textbf{Competing Methods - Input RGB Video}}\\
    \hline
     \multirow{3}{*}{Input RGB Video} 
     & Vid2CAD* \cite{vid2cad} & 38.6 & 38.0 & 22.9 & 6.2 & 81.2 & 76.6 & 69.5 & 3200\\
     & Ours (Video) & 48.2 & 52.2 & 24.7 & 3.7 & 94.7 & 79.8 & 41.7 & 100\\
     & Ours (Video, same retrieval Vid2CAD) & 48.2 & 52.9 & 29.6 & 7.7 & 95.3 & 87.7 & 71.4 & 100\\
     \hline
      \multicolumn{10}{c}{\textbf{Ablation Experiments- Input RGB-D Scan}}\\
     \hline
     \multirow{3}{*}{Embedding Distillation} & 2-step retrieval: pred bbox & 61.7 & 43.4 & 15.6 & 1.2 & 77.7 & 51.0 & 17.6 & 104\\
    & 2-step retrieval: nearest GT bbox & 61.7 & 61.7 & 30.6 & 4.1 & 93.1 & 78.1 & 36.1 & 104\\
    & Embedding distillation & 61.7 & 68.9 & 41.7 & 7.0 & 95.3 & 83.1 & 44.1 & 50 \\
     \hline
     \multirow{4}{*}{\makecell{Auxiliary Tasks for Training Encoder}}
    & Triplet & 62.3 & 67.9 & 38.3 & 5.5 & 94.8 & 81.1 & 41.1 & 50 \\
    & Tiplet + Chamfer & 61.0 & 67.5 & 38.7 & 7.3 & 95.2 & 82.0 & 42.3 & 50\\
    & Triplet + Segmentation & 61.3 & 68.8 & 41.5 & 7.5 & 95.7 & 84.3 & 43.6 & 50\\
    & Triplet + Chamfer + Segmentation & 61.7 & 68.9 & 41.7 & 7.0 & 95.3 & 83.1 & 44.1 & 50 \\
    \hline
    \multirow{2}{*}{Encoder Architecture}
     & PointNet \cite{pointnet++} & 61.5 & 61.2 & 29.6 & 4.0 & 92.0 & 74.0 & 30.9 & 50\\
     & Perceiver \cite{perceiver} & 62.3 & 67.9 & 38.3 & 5.5 & 94.8 & 81.1 & 41.1 & 50 \\
     \hline
     \multirow{2}{*}{Different Input Sources}
     & ScanNet (Gray) & 60.4 & 67.7 & 40.1 & 6.6 & 95.6 & 82.9 & 43.1 & 50\\
     & DG Recon \cite{DGrecon} (Gray) & 48.2 & 52.2 & 24.7 & 3.7 & 94.7 & 79.8 & 41.7 & 100\\
     \hline
    \multirow{5}{*}{Input Points} & 5 K & 51.4 & 56.8 & 29.8 & 4.0 & 95.3 & 81.9 & 43.0 & 42 \\
    & 10 K & 54.9 & 60.1 & 34.2 & 5.2 & 95.2 & 82.0 & 43.6 & 46 \\
    & 20 K & 57.8 & 65.6 & 37.8 & 6.3 & 95.5 & 83.3 & 44.4 & 47 \\
    & 50 K & 61.7 & 68.9 & 41.7 & 7.0 & 95.3 & 83.1 & 44.1 & 50 \\
    & 100 K & 61.8 & 69.8 & 41.7 & 7.5 & 95.5 & 83.6 & 44.0 & 50 \\
    \end{tabular}}
    \caption{\textbf{Alignment, reconstruction and shape accuracy on Scan2CAD \cite{scan2cad}} in comparison to competing methods and for various ablations. All accuracies are percentages and higher is better. Compared to Tab. 2 in the main paper, here we provide additional results for different settings of the threshold $\mu$ for computing the reconstruction and shape accuracy as explained in Sec. 4.2. Note that ScanNotate \cite{scannotate} initialises its CAD alignments from their ground-truth poses and Vid2CAD \cite{vid2cad} constraints its CAD retrieval to the very small ground-truth scene pool, making some of their results not exactly comparable to ours.}
    \label{table_supp_mat}

\end{table}
\begin{figure*}[t]
    \centering
    \includegraphics[width=1.0\linewidth]{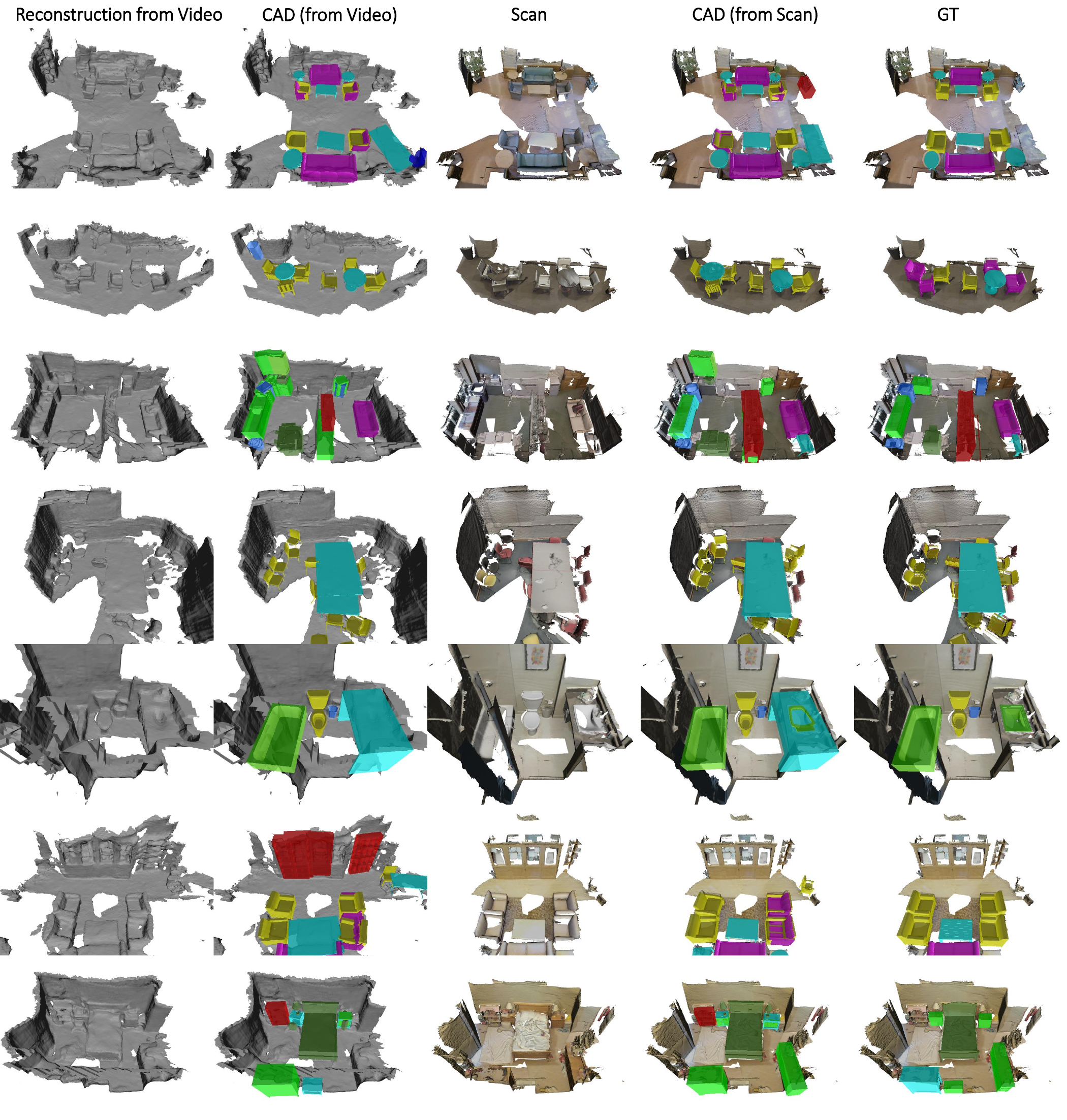}
    \caption{\textbf{Further Qualitative Visualisation on ScanNet.}
    Column 1 shows the reconstruction generated by applying \cite{DGrecon} to the input video. Column 2 shows the CAD retrieval and alignments predicted by FastCAD when operating on the reconstruction in column 1. Columns 3 and 4 show the input scan from ScanNet \cite{scannet} and the CAD alignments FastCAD predicts for it. Column 5 shows the ground-truth CAD alignments from Scan2CAD \cite{scan2cad}.
    }
    \label{fig_supp_mat_dg_recon_scannet}
\end{figure*}
\begin{figure*}[t]
    \centering
    \includegraphics[width=1.0\linewidth]{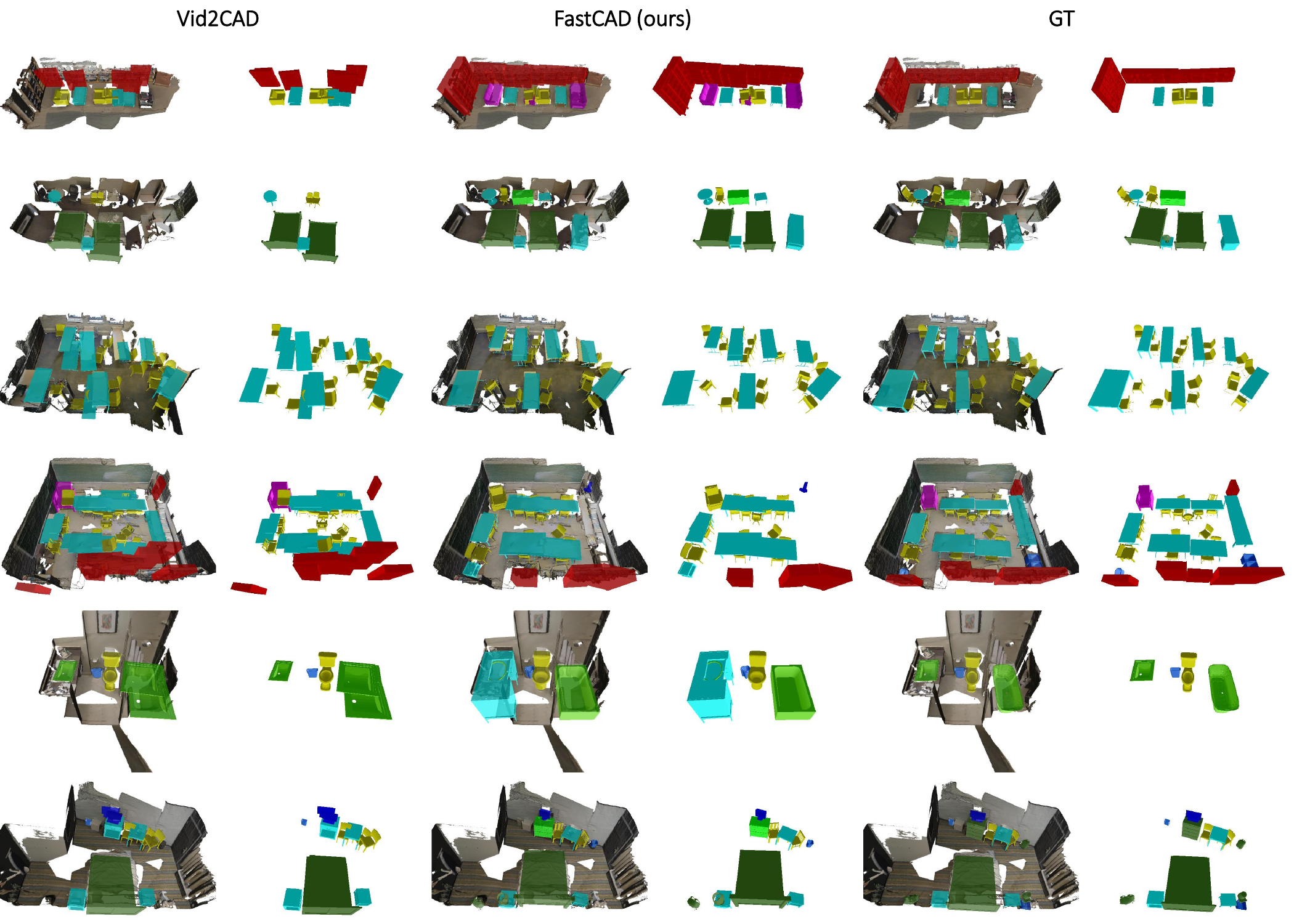}
    \caption{\textbf{Qualitative Comparison to Vid2CAD.}
    To obtain CAD alignments, we apply FastCAD to the output of \cite{DGrecon}, which uses the video of the scene as input. For a given scene, Vid2CAD \cite{vid2cad} limits its retrieval to the small ground truth scene pool, whereas we retrieve from all CAD models in the Scan2CAD \cite{scan2cad} training set.
    We find that CAD alignments produced with Vid2CAD \cite{vid2cad} are often noisy, not necessarily matching the actual object alignments in the scene. In contrast, CAD alignments produced with FastCAD are accurate, explaining the input scene well.}
    \label{fig_supp_mat_qualitative_vid2cad}
\end{figure*}

\begin{figure*}[t]
    \centering
    \includegraphics[width=1.0\linewidth]{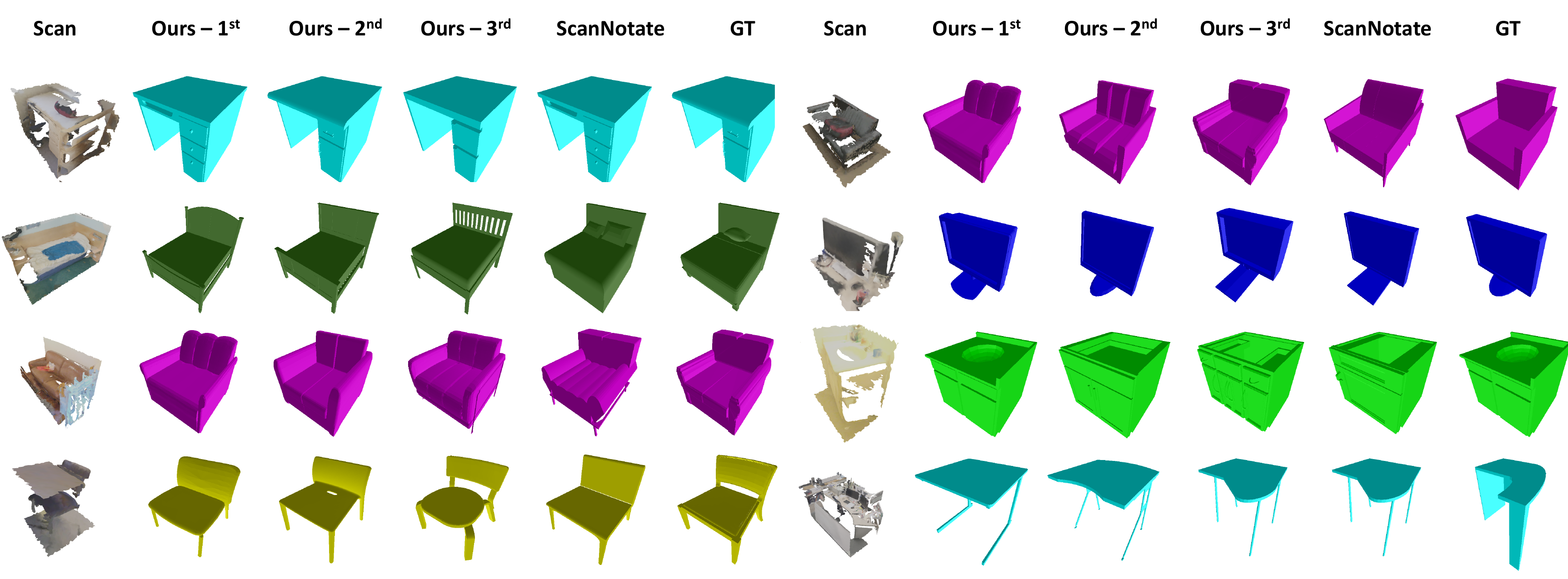}
    \caption{\textbf{Qualitative Visualisation of CAD retrievals.}
    Note that the input to FastCAD from which a shape embedding $\hat{\boldsymbol{w}}$ is predicted is the scan of the entire scene. However, for clearer visualisation, we only show the cropped part of the scan for which a CAD model is retrieved. FastCADs CAD retrievals are of high quality, in many cases as good as those obtained with the pseudo-label generation method ScanNotate \cite{scannotate} or from the annotations from Scan2CAD \cite{scan2cad} themselves.
    }
    \label{fig_supp_mat_retrieval}
\end{figure*}

\begin{figure*}[t]
    \centering
    \includegraphics[width=1.0\linewidth]{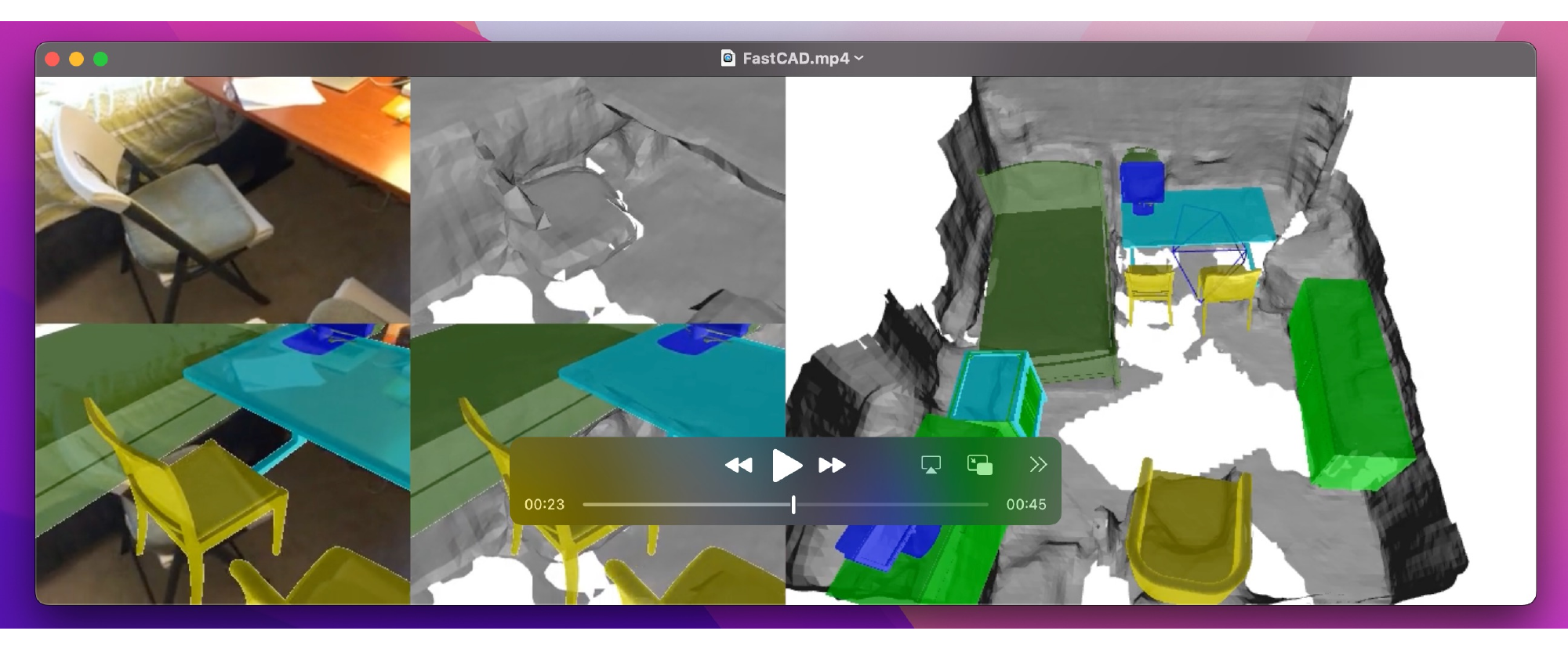}
    \caption{\textbf{Visualisation Video}. We visualise our online CAD retrieval and alignments in a visualisation video. The top left images show the input RGB video and the reconstruction generated with \cite{DGrecon}. The bottom left images show the aligned CAD models overlayed to the input video and the reconstruction from \cite{DGrecon}. The image on the right shows the global scene reconstruction and current camera pose. One can see that FastCAD is able to create accurate CAD-based scene reconstructions for a diverse set of input scenes. The full video is available
\href{https://youtu.be/KIcFRBu0CiE}{here}.}
\label{fig_supp_mat_visualisation_video}
\end{figure*}

We include extra qualitative visualisations. Fig. \ref{fig_supp_mat_dg_recon_scannet} provides additional visualisations of FastCAD when using either the output of \cite{DGrecon} or directly the scans from ScanNet \cite{scannet} as input. While CAD alignments are of high quality in both cases, alignments from scans are consistently more accurate. This is because the reconstructions generated with \cite{DGrecon} can be noisy or miss crucial details which poses a challenge for the subsequent CAD retrieval and alignment.

Fig. \ref{fig_supp_mat_qualitative_vid2cad} shows a qualitative comparison of our method compared to Vid2CAD \cite{vid2cad}. It can be seen that FastCAD is considerably more accurate in its CAD alignments. We believe that this is largely due to the design decision to perform CAD alignment in 3D as opposed to relying on detecting objects in 2D and matching them across frames. Inaccuracies in 2D detection and wrong associations across frames are likely the cause for many of the misalignments of Vid2CAD observed in Fig \ref{fig_supp_mat_qualitative_vid2cad}.

We include additional visualisations for FastCADs CAD retrieval in Fig. \ref{fig_supp_mat_retrieval}. In general, the retrieved shapes match the underlying objects very well (e.g. the retrieved tables in the first row). However, for some objects the retrieved CAD models are not closely fitting. We find this is the case particularly for objects with missing scene geometry (e.g. the third CAD retrieval for the chair in the last row) or for objects with a lot of clutter (e.g. the retrieved tables in the last row). 

Finally, we provide a visualisation video (screenshot in Fig. \ref{fig_supp_mat_visualisation_video}, full video available \hyperlink{https://www.youtube.com/watch?v=KIcFRBu0CiE}{here}.) showcasing FastCAD's ability to perform accurate CAD-based reconstructions from videos online. From the start of the sequence, the aligned CAD models provide a faithful reconstruction of the underlying scene. Failure modes can include overlapping CAD models in the reconstruction when partially seen objects are revealed further as well as sub-optimal shape retrieval for certain objects. 

\clearpage
\bibliographystyle{plainnat}
\bibliography{main}

\begin{thebibliography}{39}
\providecommand{\natexlab}[1]{#1}
\providecommand{\url}[1]{\texttt{#1}}
\expandafter\ifx\csname urlstyle\endcsname\relax
  \providecommand{\doi}[1]{doi: #1}\else
  \providecommand{\doi}{doi: \begingroup \urlstyle{rm}\Url}\fi

\bibitem[Ainetter et~al.(2023)Ainetter, Stekovic, Fraundorfer, and Lepetit]{scannotate}
Stefan Ainetter, Sinisa Stekovic, Friedrich Fraundorfer, and Vincent Lepetit.
\newblock Automatically annotating indoor images with cad models via rgb-d scans.
\newblock In \emph{IEEE/CVF Winter Conf. App. Comput. Vis.}, 2023.

\bibitem[Avetisyan et~al.(2019{\natexlab{a}})Avetisyan, Dahnert, Dai, Savva, Chang, and Niessner]{scan2cad}
Armen Avetisyan, Manuel Dahnert, Angela Dai, Manolis Savva, Angel~X. Chang, and Matthias Niessner.
\newblock Scan2cad: Learning cad model alignment in rgb-d scans.
\newblock In \emph{IEEE Conf. Comput. Vis. Pattern Recog.}, 2019{\natexlab{a}}.

\bibitem[Avetisyan et~al.(2019{\natexlab{b}})Avetisyan, Dai, and Niessner]{end_to_end_cad_retrieval}
Armen Avetisyan, Angela Dai, and Matthias Niessner.
\newblock End-to-end cad model retrieval and 9dof alignment in 3d scans.
\newblock In \emph{Int. Conf. Comput. Vis.}, 2019{\natexlab{b}}.

\bibitem[Avetisyan et~al.(2020)Avetisyan, Khanova, Choy, Dash, Dai, and Nießner]{scenecad}
Armen Avetisyan, Tatiana Khanova, Christopher Choy, Denver Dash, Angela Dai, and Matthias Nießner.
\newblock Scenecad: Predicting object alignments and layouts in rgb-d scans.
\newblock In \emph{Eur. Conf. Comput. Vis.}, 2020.

\bibitem[Bozic et~al.(2021)Bozic, Palafox, Thies, Dai, and Nie{\ss}ner]{transformerfusion}
Aljaz Bozic, Pablo Palafox, Justus Thies, Angela Dai, and Matthias Nie{\ss}ner.
\newblock Transformerfusion: Monocular rgb scene reconstruction using transformers.
\newblock \emph{Adv. Neural Inform. Process. Syst.}, 2021.

\bibitem[Chang et~al.(2015)Chang, Funkhouser, Guibas, Hanrahan, Huang, Li, Savarese, Savva, Song, Su, et~al.]{shapenet}
Angel~X Chang, Thomas Funkhouser, Leonidas Guibas, Pat Hanrahan, Qixing Huang, Zimo Li, Silvio Savarese, Manolis Savva, Shuran Song, Hao Su, et~al.
\newblock Shapenet: An information-rich 3d model repository.
\newblock \emph{arXiv preprint arXiv:1512.03012}, 2015.

\bibitem[Cheng et~al.(2021)Cheng, Sheng, Shi, Yang, and Xu]{brnet}
Bowen Cheng, Lu~Sheng, Shaoshuai Shi, Ming Yang, and Dong Xu.
\newblock Back-tracing representative points for voting-based 3d object detection in point clouds.
\newblock In \emph{IEEE Conf. Comput. Vis. Pattern Recog.}, 2021.

\bibitem[Choy et~al.(2019)Choy, Gwak, and Savarese]{minkowski_1}
Christopher Choy, JunYoung Gwak, and Silvio Savarese.
\newblock 4d spatio-temporal convnets: Minkowski convolutional neural networks.
\newblock In \emph{IEEE Conf. Comput. Vis. Pattern Recog.}, 2019.

\bibitem[Choy et~al.(2020)Choy, Dong, and Koltun]{procrustes}
Christopher Choy, Wei Dong, and Vladlen Koltun.
\newblock Deep global registration.
\newblock In \emph{IEEE Conf. Comput. Vis. Pattern Recog.}, 2020.

\bibitem[Contributors(2020)]{mmdet3d2020}
MMDetection3D Contributors.
\newblock {MMDetection3D: OpenMMLab} next-generation platform for general {3D} object detection, 2020.

\bibitem[Dahnert et~al.(2019)Dahnert, Dai, Guibas, and Nie{\ss}ner]{joint_embedding}
Manuel Dahnert, Angela Dai, Leonidas Guibas, and Matthias Nie{\ss}ner.
\newblock Joint embedding of 3d scan and cad objects.
\newblock In \emph{Int. Conf. Comput. Vis.}, 2019.

\bibitem[Dai et~al.(2017)Dai, Chang, Savva, Halber, Funkhouser, and Nießner]{scannet}
Angela Dai, Angel~X. Chang, Manolis Savva, Maciej Halber, Thomas Funkhouser, and Matthias Nießner.
\newblock Scannet: Richly-annotated 3d reconstructions of indoor scenes.
\newblock In \emph{IEEE Conf. Comput. Vis. Pattern Recog.}, 2017.

\bibitem[Gkioxari et~al.(2019)Gkioxari, Malik, and Johnson]{meshrcnn}
Georgia Gkioxari, Jitendra Malik, and Justin Johnson.
\newblock Mesh r-cnn.
\newblock In \emph{Int. Conf. Comput. Vis.}, 2019.

\bibitem[Gwak et~al.(2020{\natexlab{a}})Gwak, Choy, and Savarese]{gsdn}
JunYoung Gwak, Christopher~B Choy, and Silvio Savarese.
\newblock Generative sparse detection networks for 3d single-shot object detection.
\newblock In \emph{Eur. Conf. Comput. Vis.}, 2020{\natexlab{a}}.

\bibitem[Gwak et~al.(2020{\natexlab{b}})Gwak, Choy, and Savarese]{minkowski_2}
JunYoung Gwak, Christopher~B Choy, and Silvio Savarese.
\newblock Generative sparse detection networks for 3d single-shot object detection.
\newblock In \emph{Eur. Conf. Comput. Vis.}, 2020{\natexlab{b}}.

\bibitem[Hampali et~al.(2021)Hampali, Stekovic, Sarkar, Kumar, Fraundorfer, and Lepetit]{mcss}
Shreyas Hampali, Sinisa Stekovic, Sayan~Deb Sarkar, Chetan~Srinivasa Kumar, Friedrich Fraundorfer, and Vincent Lepetit.
\newblock Monte carlo scene search for 3d scene understanding.
\newblock In \emph{IEEE Conf. Comput. Vis. Pattern Recog.}, 2021.

\bibitem[Jaegle et~al.(2021)Jaegle, Gimeno, Brock, Zisserman, Vinyals, and Carreira]{perceiver}
Andrew Jaegle, Felix Gimeno, Andrew Brock, Andrew Zisserman, Oriol Vinyals, and Joao Carreira.
\newblock Perceiver: General perception with iterative attention.
\newblock In \emph{Int. Conf. Mach. Learn.}, 2021.

\bibitem[Ju et~al.(2023)Ju, Tseng, Bailo, Dikov, and Ghafoorian]{DGrecon}
Jihong Ju, Ching~Wei Tseng, Oleksandr Bailo, Georgi Dikov, and Mohsen Ghafoorian.
\newblock Dg-recon: Depth-guided neural 3d scene reconstruction.
\newblock In \emph{Int. Conf. Comput. Vis.}, 2023.

\bibitem[Kuo et~al.(2020)Kuo, Angelova, Lin, and Dai]{mask2cad}
Weicheng Kuo, Anelia Angelova, Tsung-Yi Lin, and Angela Dai.
\newblock Mask2cad: 3d shape prediction by learning to segment and retrieve.
\newblock In \emph{Eur. Conf. Comput. Vis.}, 2020.

\bibitem[Kuo et~al.(2021)Kuo, Angelova, Lin, and Dai]{patch2cad}
Weicheng Kuo, Anelia Angelova, Tsung-Yi Lin, and Angela Dai.
\newblock Patch2cad: Patchwise embedding learning for in-the-wild shape retrieval from a single image.
\newblock \emph{Int. Conf. Comput. Vis.}, 2021.

\bibitem[Langer et~al.(2021)Langer, Budvytis, and Cipolla]{leveraging_geometry}
Florian Langer, Ignas Budvytis, and Roberto Cipolla.
\newblock Leveraging geometry for shape estimation from a single rgb image.
\newblock In \emph{Brit. Mach. Vis. Conf.}, 2021.

\bibitem[Li et~al.(2021)Li, DeTone, Chen, Vo, Reid, Rezatofighi, Sweeney, Straub, and Newcombe]{odam}
Kejie Li, Daniel DeTone, Steven Chen, Minh Vo, Ian Reid, Hamid Rezatofighi, Chris Sweeney, Julian Straub, and Richard Newcombe.
\newblock Odam: Object detection, association, and mapping using posed rgb video.
\newblock In \emph{Int. Conf. Comput. Vis.}, 2021.

\bibitem[Li et~al.(2015)Li, Su, Qi, Fish, Cohen-Or, and Guibas]{image_purification}
Yangyan Li, Hao Su, Charles~Ruizhongtai Qi, Noa Fish, Daniel Cohen-Or, and Leonidas~J. Guibas.
\newblock Joint embeddings of shapes and images via cnn image purification.
\newblock \emph{ACM Trans. Graph.}, 2015.

\bibitem[Liu et~al.(2021)Liu, Zhang, Cao, Hu, and Tong]{groupfree}
Ze~Liu, Zheng Zhang, Yue Cao, Han Hu, and Xin Tong.
\newblock Group-free 3d object detection via transformers.
\newblock In \emph{Int. Conf. Comput. Vis.}, 2021.

\bibitem[Loshchilov and Hutter(2019)]{adamw}
Ilya Loshchilov and Frank Hutter.
\newblock Decoupled weight decay regularization.
\newblock In \emph{Int. Conf. Learn. Represent.}, 2019.

\bibitem[Maninis et~al.(2022)Maninis, Popov, Nießner, and Ferrari]{vid2cad}
Kevis-Kokitsi Maninis, Stefan Popov, Matthias Nießner, and Vittorio Ferrari.
\newblock Vid2cad: Cad model alignment using multi-view constraints from videos.
\newblock \emph{IEEE Transactions on Pattern Analysis and Machine Inttelligence}, 2022.

\bibitem[Misra et~al.(2021)Misra, Girdhar, and Joulin]{3detr}
Ishan Misra, Rohit Girdhar, and Armand Joulin.
\newblock {An End-to-End Transformer Model for 3D Object Detection}.
\newblock In \emph{Int. Conf. Comput. Vis.}, 2021.

\bibitem[Pham et~al.(2018)Pham, Tran, Li, Xiang, Zhou, Nie, Liu, Su, Tran, Bui, Do, Ninh, Le, Dao, Nguyen, Do, Duong, Hua, Yu, Nguyen, and Yeung]{shrec}
Quang-Hieu Pham, Minh-Khoi Tran, Wenhui Li, Shu Xiang, Heyu Zhou, Weizhi Nie, Anan Liu, Yuting Su, Minh-Triet Tran, Ngoc-Minh Bui, Trong-Le Do, Tu~V. Ninh, Tu-Khiem Le, Anh-Vu Dao, Vinh-Tiep Nguyen, Minh~N. Do, Anh-Duc Duong, Binh-Son Hua, Lap-Fai Yu, Duc~Thanh Nguyen, and Sai-Kit Yeung.
\newblock {RGB-D Object-to-CAD Retrieval}.
\newblock In \emph{Eurographics Workshop on 3D Object Retrieval}, 2018.

\bibitem[Qi et~al.(2019)Qi, Litany, He, and Guibas]{votenet}
Charles~R Qi, Or~Litany, Kaiming He, and Leonidas~J Guibas.
\newblock Deep hough voting for 3d object detection in point clouds.
\newblock In \emph{Int. Conf. Comput. Vis.}, 2019.

\bibitem[Qi et~al.(2017)Qi, Yi, Su, and Guibas]{pointnet++}
Charles~Ruizhongtai Qi, Li~Yi, Hao Su, and Leonidas~J Guibas.
\newblock Pointnet++: Deep hierarchical feature learning on point sets in a metric space.
\newblock \emph{Adv. Neural Inform. Process. Syst.}, 2017.

\bibitem[Rukhovich et~al.(2022)Rukhovich, Vorontsova, and Konushin]{fcaf3d}
Danila Rukhovich, Anna Vorontsova, and Anton Konushin.
\newblock Fcaf3d: fully convolutional anchor-free 3d object detection.
\newblock In \emph{Eur. Conf. Comput. Vis.}, 2022.

\bibitem[Rukhovich et~al.(2023)Rukhovich, Vorontsova, and Konushin]{tr3d}
Danila Rukhovich, Anna Vorontsova, and Anton Konushin.
\newblock Tr3d: Towards real-time indoor 3d object detection.
\newblock In \emph{IEEE Int. Conf. Image Process.}, 2023.

\bibitem[Sayed et~al.(2022)Sayed, Gibson, Watson, Prisacariu, Firman, and Godard]{simplerecon}
Mohamed Sayed, John Gibson, Jamie Watson, Victor Prisacariu, Michael Firman, and Cl{\'e}ment Godard.
\newblock Simplerecon: 3d reconstruction without 3d convolutions.
\newblock In \emph{Eur. Conf. Comput. Vis.}, 2022.

\bibitem[Schroff et~al.(2015)Schroff, Kalenichenko, and Philbin]{triplet}
Florian Schroff, Dmitry Kalenichenko, and James Philbin.
\newblock Facenet: A unified embedding for face recognition and clustering.
\newblock In \emph{IEEE Conf. Comput. Vis. Pattern Recog.}, 2015.

\bibitem[Tyszkiewicz et~al.(2022)Tyszkiewicz, Maninis, Popov, and Ferrari]{raytran}
Michał~J. Tyszkiewicz, Kevis-Kokitsi Maninis, Stefan Popov, and Vittorio Ferrari.
\newblock Raytran: 3d pose estimation and shape reconstruction of multiple objects from videos with ray-traced transformers.
\newblock In \emph{Eur. Conf. Comput. Vis.}, 2022.

\bibitem[Wang et~al.(2022)Wang, Shi, Yang, Fang, Qian, Li, Schiele, and Wang]{rbgnet}
Haiyang Wang, Shaoshuai Shi, Ze~Yang, Rongyao Fang, Qi~Qian, Hongsheng Li, Bernt Schiele, and Liwei Wang.
\newblock Rbgnet: Ray-based grouping for 3d object detection.
\newblock In \emph{IEEE Conf. Comput. Vis. Pattern Recog.}, 2022.

\bibitem[You et~al.(2020)You, Li, Hseu, Song, Demmel, and Hsieh]{lamb}
Yang You, Jing Li, Jonathan Hseu, Xiaodan Song, James Demmel, and Cho{-}Jui Hsieh.
\newblock Reducing {BERT} pre-training time from 3 days to 76 minutes.
\newblock \emph{Int. Conf. Learn. Represent.}, 2020.

\bibitem[Zhang et~al.(2020)Zhang, Sun, Yang, and Huang]{h3dnet}
Zaiwei Zhang, Bo~Sun, Haitao Yang, and Qi-Xing Huang.
\newblock H3dnet: 3d object detection using hybrid geometric primitives.
\newblock In \emph{Eur. Conf. Comput. Vis.}, 2020.

\bibitem[Zheng et~al.(2020)Zheng, Wang, Liu, Li, Ye, and Ren]{diou}
Zhaohui Zheng, Ping Wang, Wei Liu, Jinze Li, Rongguang Ye, and Dongwei Ren.
\newblock Distance-iou loss: Faster and better learning for bounding box regression.
\newblock In \emph{AAAI}, 2020.

\end{thebibliography}

\end{document}